\documentclass{article}
\usepackage[table]{xcolor}
\usepackage{spconf}
\usepackage{amsmath,amsfonts}
\usepackage{algorithmic}
\usepackage{algorithm}
\usepackage{array}
\usepackage[caption=false,font=normalsize,labelfont=sf,textfont=sf]{subfig}
\usepackage{textcomp}
\usepackage{stfloats}
\usepackage{url}
\usepackage{verbatim}
\usepackage{graphicx}
\usepackage{cite}
\usepackage{color}
\usepackage{physics}
\usepackage{comment}
\usepackage{pifont}
\usepackage{xcolor}
\usepackage{arydshln}
\usepackage{booktabs}
\usepackage{multirow}
\usepackage{subcaption}
\usepackage{mdwlist}
\usepackage{hyperref}
\usepackage[capitalize]{cleveref}
\usepackage{enumitem}

\definecolor{Gray}{gray}{0.9}
\newcommand{\softmax}{\sigma}
\newcommand{\ignore}[1]{}

\title{IDAG-Edit: Multi-Object Video Editing via \underline{I}nstance-\underline{D}ecoupled \underline{A}ttention and \underline{G}uidance}
\name{Yuan-Zhih Lin$^{1}$ \qquad Huu-Thang Nguyen$^{1}$ \qquad Huu-Phu Do$^{1}$ \qquad Hong-Han Shuai$^{1}$ \qquad Ching-Chun Huang\sthanks{Corresponding author.}$^{1}$}
\address{$^{1}$Department of Computer Science, National Yang Ming Chiao Tung University, Taiwan}
\begin{document}
%
\maketitle
\begin{abstract}
%
Diffusion-based video editing has made significant progress; however, achieving precise and temporally consistent object-level control, especially in multi-object scenarios, remains challenging due to attention leakage, identity drift, and unstable temporal dynamics. In this work, we propose IDAG-Edit, a training-free framework for fine-grained multi-object video editing with strong temporal consistency. The framework adopts \textit{Layout-guided Attention Modulation} to facilitate coherent multi-object editing, while \textit{Instance-level Masks} are introduced to preserve individual object identity and enforce localized attention within each object region, thereby enabling fine-grained, object-level editing. 
Extensive qualitative and quantitative evaluations demonstrate that our method improves temporal stability and multi-object controllability over state-of-the-art video editing approaches.
\end{abstract}
\section{Introduction}
\label{sec:intro}

Video editing aims to modify content according to textual prompts while preserving visual realism and temporal coherence across frames. Despite rapid advances in diffusion-based video editing methods, achieving precise and temporally consistent object-level control—particularly in scenarios involving multiple objects and attributes—remains challenging.

Existing video editing methods~\cite{flatten, flowvid, tuneavideo, controlvideo, videograin, gav, fatezero, pix2video, tokenflow} build on pretrained text-to-image (T2I) diffusion models~\cite{stablediffusion} and extend them to video using auxiliary mechanisms, including optical-flow–based feature alignment~\cite{flatten, flowvid}, cross-frame attention regularization~\cite{tuneavideo, controlvideo}, and feature propagation along inversion trajectories~\cite{tokenflow}. More recent approaches explore multi-object editing through explicit spatial guidance: Ground-A-Video~\cite{gav} uses text–bounding box grounding but suffers from attention leakage under overlapping boxes, while VideoGrain~\cite{videograin} mitigates this issue via mask-aware attention modulation. However, temporal dependencies are imposed externally rather than learned within the diffusion process, resulting in only approximate temporal coherence and causing flickering, trajectory drift, and identity inconsistency.

\begin{figure}[t]
    \includegraphics[width=1.0\linewidth]{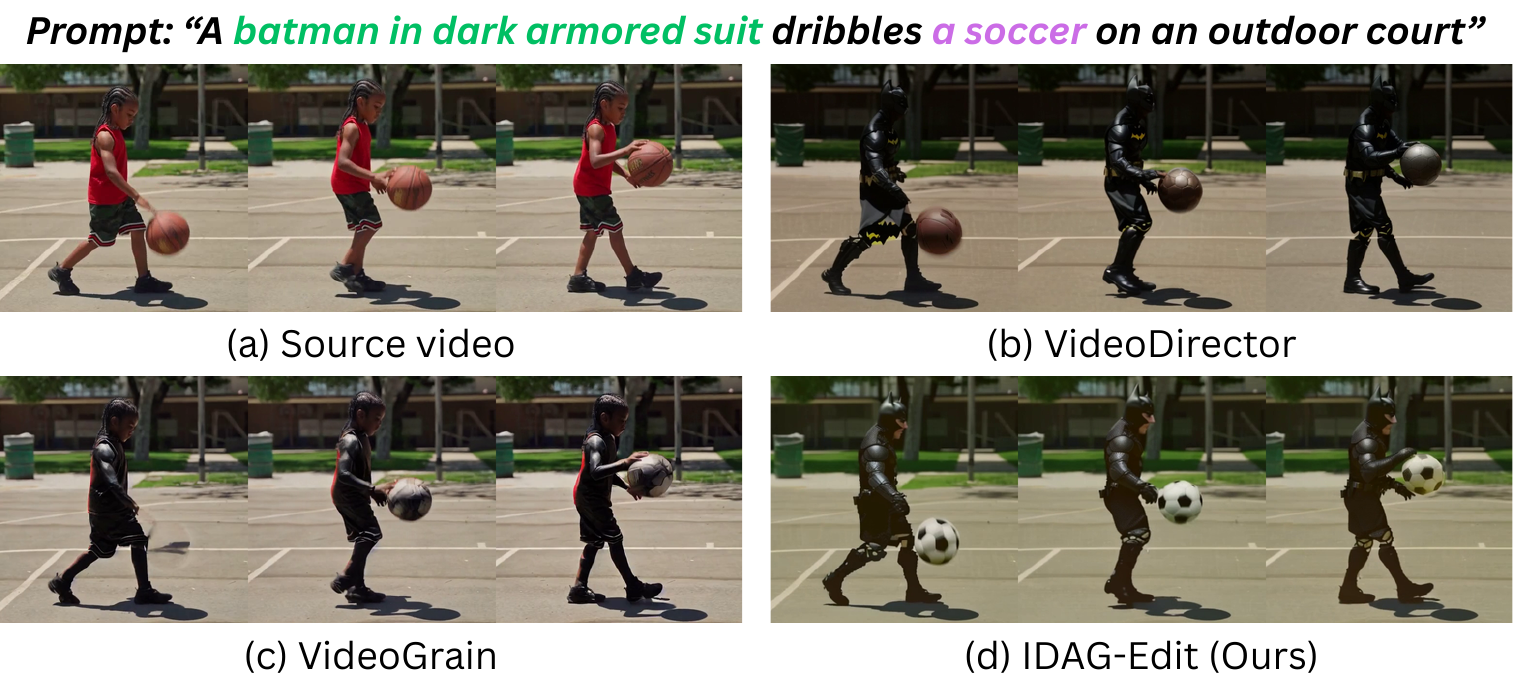}
    \caption{\textbf{Comparison with baseline methods.} VideoDirector~\cite{videodirector} struggles with multi-object editing and VideoGrain~\cite{videograin} lacks temporal consistency. Our method achieves both accurate editing and superior temporal stability.}
    \label{fig:fig1}
    \vspace{-2mm}
\end{figure}

In parallel, recent video diffusion models have been increasingly adopted for video editing by modeling spatio-temporal dynamics to improve temporal coherence.
Methods such as AnyV2V~\cite{anyv2v} and StableV2V~\cite{stablev2v} follow image-to-video diffusion frameworks, editing the first frame and propagating edits to subsequent frames via temporally conditioned attention, making their performance sensitive to the initial frame quality. VideoDirector~\cite{videodirector} instead leverages AnimateDiff~\cite{animatediff} as a text-to-video (T2V) backbone and introduces spatial–temporal decoupled guidance to align editing and reconstruction attention, together with spatial masks for localized control. However, these methods primarily target single-object or scene-level edits and do not explicitly disentangle object-level semantics, often causing attribute entanglement, unintended modifications and identity drift in multi-object scenarios as shown in Fig. \ref{fig:fig1}.



To address these limitations, we present IDAG-Edit, a framework for fine-grained and temporally coherent multi-object video editing, characterized by two core contributions. First, \textit{Layout-Guided Cross-Attention Control} (LG-CA) is incorporated to enforce text–spatial consistency across regions, enabling coherent multi-object editing. Second, \textit{Instance-level Masks} is presented to support fine-grained, object-level editing. Specifically, \textit{Instance-Aware Spatio-Temporal Decoupled Guidance} (IA-STDG) is proposed to alleviate attribute entanglement caused by shared foreground guidance in multi-object scenes, thereby preserving individual object identity and motion trajectories. During editing, \textit{Instance-Decoupled Self-Attention} (ID-SA) is further introduced to constrains self-attention aggregation within each object region, ensuring localized feature interactions, reducing cross-object feature bleeding, and promoting sharper object boundaries.
In summary, our main contributions are as follow:
\begin{itemize}[leftmargin=10pt, parsep=0pt]
    \item We propose \textbf{IDAG-Edit}, a training-free video diffusion editing framework for fine-grained multi-object editing with strong temporal consistency.
   \item We adopt LG-CA to facilitate coherent multi-object editing, and propose instance-level masks via IA-STDG and ID-SA to enable fine-grained, object-level control.
   
    \item Extensive experiments demonstrate that IDAG-Edit outperforms state-of-the-art approaches in temporal stability and multi-object controllability.
\end{itemize}


\begin{figure*}[t]
    \centering
    \includegraphics[width=0.95\linewidth]{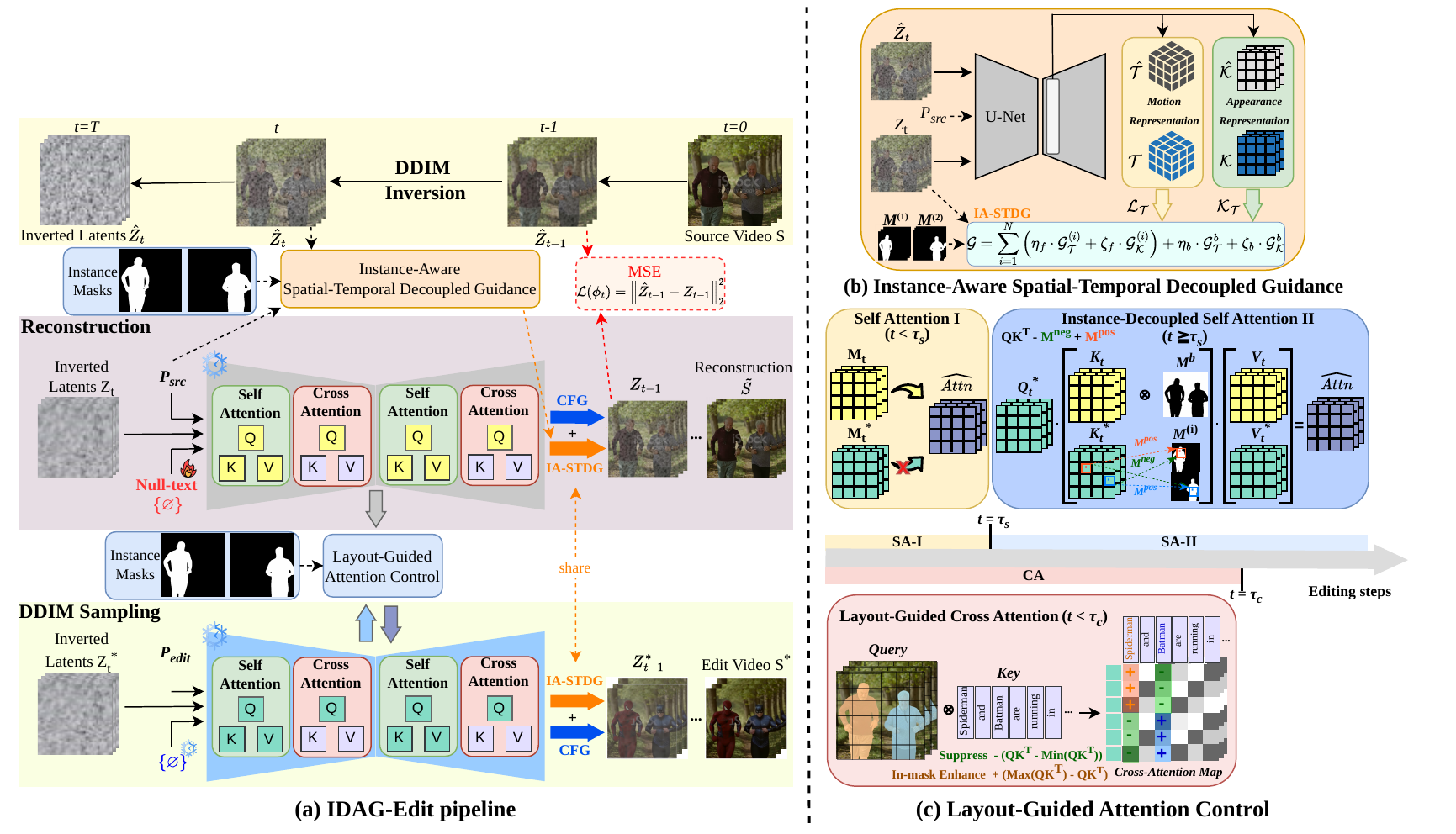}

    \caption{\textbf{Overview of the IDAG-Edit pipeline}. (a) The framework operates in three stages: DDIM inversion and Reconstruction are first performed to store guidance parameters, followed by an editing phase (DDIM sampling) that preserves source structure while applying modifications. (b) IA-STDG: This guidance steers the denoising trajectory with multi-grained control for precise object-level decoupling.
    (c) Attention Control: Following~\cite{videodirector}, we apply Self-Attention I ($t < \tau_s$), ID-SA ($t \geq \tau_s$), and LG-CA ($t < \tau_c$). Specifically, LG-CA regulates spatial weights, while ID-SA constrains queries to target regions to prevent attribute leakage and ensure feature separation.}

    \label{fig:overview}
\end{figure*}
\section{Methodology}
\label{sec:Methodology}
\subsection{Overview Framework}
\label{ssec:Overview}



As shown in \cref{fig:overview}, building upon VideoDirector~\cite{videodirector}, we formulate video editing as a three-stage pipeline that decouples content preservation from targeted manipulation. Given a source video $S$, source prompt $P_{\text{src}}$, instance-forground editing masks $\mathcal{M}^{(i)}$ and  background mask $\mathcal{M}^{(b)}$, faithful video reconstruction $\tilde{S}$ is achieved via DDIM inversion~\cite{ddim} and a denoising process guided by multi-frame null-text optimization and Instance-Aware Spatial-Temporal Decoupled Guidance (IA-STDG), which preserves individual object identity and motion trajectories (\Cref{ssec:IA-STDG}). Editing is subsequently carried out during the sampling stage, conditioned on an edited prompt $P_{\text{edit}}$ and the instance masks. Within this editing phase, \textit{Layout-Guided Cross-Attention Control} (LG-CA) is adopted to enforce spatial consistency across regions (\Cref{ssec:LGCA}), while \textit{Instance-Decoupled Self-Attention} (ID-SA) is employed to constrain self-attention within each object region, enabling localized feature interactions (\Cref{ssec:ID-SA}). 

\vspace{-7pt}
\subsection{Instance-Aware Spatial-Temporal Decoupled Guidance (IA-STDG)}
\label{ssec:IA-STDG}

Spatial–Temporal Decoupled Guidance (STDG)~\cite{videodirector} has been shown to effectively preserve motion and appearance consistency in single-object video editing. The mechanism calculates temporal ($\mathcal{L}_{T}$) and spatial loss terms ($\mathcal{L}_{K}$) by comparing attention maps and keys between pivotal inversion and denoising stages. The corresponding guidance is then derived from the gradients of these losses with respect to the noisy latent at each timestep. However, this global strategy is less effective in multi-object scenes. When multiple distinct instances share a global gradient field, their optimization trajectories can interfere with one another, which entangles their independent motions and attributes, eventually leading to visual mixing and attribute leakage. 

To address these limitations, we introduce Instance-Aware STDG (IA-STDG), which leverages per-instance masks to compute object-specific gradients, as illustrated in Fig. \ref{fig:overview}b. Building upon the core loss formulation of STDG, we compute temporal and spatial losses independently for each foreground instance. By isolating objects within individual masks $\mathcal{M}^{(i)}$, we generate instance-specific guidance that prevents cross-object interference during the backward pass. The background guidance is similarly computed using $\mathcal{M}^{b} = 1 - \bigcup_i \mathcal{M}^{(i)}$. The overall guidance $\mathcal{G}$ at each step is formulated as a weighted sum of the instance-level and background components:
\begin{equation}
    \begin{aligned}
        \mathcal G=\sum_{i=1}^{N}\left ( \eta _f \cdot \mathcal G_{\mathcal T}^{(i)} + \zeta _f \cdot \mathcal G_{\mathcal K}^{(i)} \right )+ \eta _b \cdot \mathcal G_{\mathcal T}^{b} + \zeta _b \cdot \mathcal G_{\mathcal K}^{b}
        \end{aligned}
    \label{eq: total_guidance}
\end{equation}
where $\eta_f$, $\zeta_f$, $\eta_b$, and $\zeta_b$ are weighting coefficients for instance-level and background guidance, respectively. Finally, we integrate IA-STDG with classifier-free guidance~\cite{cfg} as an auxiliary guidance term~\cite{clsguidance} to refine the denoising trajectory.

\begin{figure*}[t]
    \centering
    \includegraphics[width=1.0\linewidth]{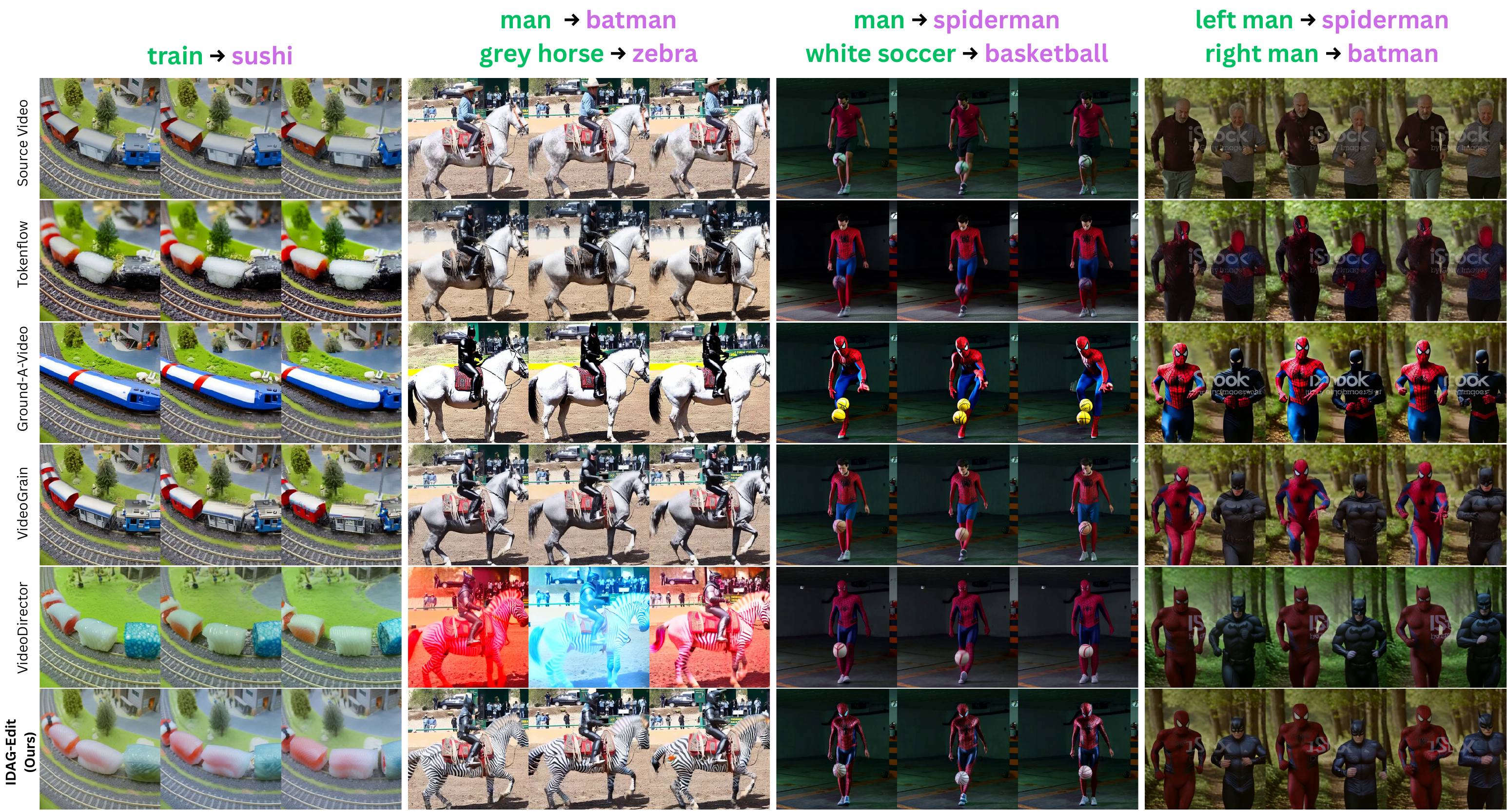}
    \caption{\textbf{Qualitative comparison.} Our method demonstrates superior editing quality over other approaches, producing more accurate and visually coherent results with improved temporal consistency.}
    \label{fig:qualitative}
\end{figure*}

\vspace{-7pt}
\subsection{Layout-Guided Cross-Attention}
\label{ssec:LGCA}
Following DenseDiffusion~\cite{densetext}, we adopt layout-guided cross-attention to align prompt semantics with their corresponding spatial regions. Specifically, during the first $\tau_c$ denoising steps, the cross-attention maps in the editing branch are modulated using the provided layout. Let $\mathcal{A}^{\text{edit}}_t$ denote the original cross-attention at timestep $t$, and the modulated attention $\widehat{\text{Attn}}_t$ is computed as:
\begin{equation}
\widehat{\text{Attn}}_t^{\text{cross}}= 
\begin{cases}
\softmax\left(\frac {\mathbf{Q}_t \mathbf{K}_t^\top + \lambda \mathbf{M}^{\text{cross}}_t}{\surd d}\right), & \text{if } t < \tau_c, \\
\mathcal{A}_t^{\text{edit}}, & \text{otherwise}.
\end{cases}
\label{eq: cross-attention}
\end{equation}
Where $\softmax(\cdot)$ denotes the softmax function applied over attention logits, and $\lambda = \xi(t) \cdot (1 - S_i)$ serves as a regularization factor, with $\xi(t)$ controlling the modulation strength over time. The modulation term $\mathbf{M}^{\text{cross}}_t \in \mathbb{R}^{F \times (H\times W) \times L}$ at denoising timestep $t$ is constructed as:
\begin{equation}
\begin{aligned}
\mathbf{M}^{\text{cross}}_t = \mathbf{C} \odot \mathbf{M}^{\text{pos}}_t - (1 - \mathbf{C}_t) \odot \mathbf{M}^{\text{neg}}_t
\end{aligned}
\label{eq:ca-bias}
\end{equation}
\begin{equation}
\begin{aligned}
\mathbf{M}^{\text{pos}}_t = \max(\mathbf{Q}_t \mathbf{K}_t^\top) - \mathbf{Q}_t \mathbf{K}_t^\top, \\
\mathbf{M}^{\text{neg}}_t = \mathbf{Q}_t \mathbf{K}_t^\top - \min(\mathbf{Q}_t \mathbf{K}_t^\top),
\end{aligned}
\label{eq:ca-max-min}
\end{equation}
where $\mathbf{M}^{\text{pos}/\text{neg}}_t$ is derived based on the difference between the original attention scores and their respective maximum or minimum values, ensuring that the modulated scores remain within a bounded range. The query-key pair condition map $\mathbf{C} \in \mathbb{R}^{F \times (H \times W) \times L}$ serves as a modulation signal that selectively adjusts attention scores to enforce text-to-region control. Formally, the condition map is constructed as:
\begin{equation}
\mathbf{C}[q, k] = 
\begin{cases}
\mathcal{M}_{i,j}, & \text{if } k \in \tau_j, \\
0, & \text{otherwise}.
\end{cases}
\end{equation}
where $q$ and $k$ denote the query and key indices, and $\tau_j$ is the set of text token indices assigned to region $j$. $\mathcal{M}_{i,j}$ is derived by broadcasting the region mask $i$ to match the corresponding text key embeddings $\mathbf{K}_{\tau_j}$ for regularization.

\vspace{-7pt}
\subsection{Instance-Decoupled Self-Attention}
\label{ssec:ID-SA}


To mitigate feature entanglement across multiple objects, we introduce Instance-Decoupled Self-Attention, where the self-attention keys and values from both reconstruction and editing branches are concatenated as $\mathbf{K}_t^{\prime} = [\mathbf{K}_t^* | \mathbf{K}_t]$ and $\mathbf{V}_t^{\prime} = [\mathbf{V}_t^* | \mathbf{V}_t] \in \mathbb{R}^{F \times (2 \cdot H \cdot W) \times C}$. Next, the attention maps are computed using queries from the editing branch against keys $\mathbf{K}_t^{\prime}$, where the interaction is partitioned: responses to editing branch keys $\mathbf{K}_t^*$ are modulated by a reweighting function $f(\cdot)$, while those to reconstruction branch keys $\mathbf{K}_t$ are constrained by mask $\mathcal{M}^b$ to prevent original content in the regions to be edited. Specifically, we modulate attentions during different timesteps as follows:
{\small
\begin{equation}
\widehat{\text{Attn}}_t^{\text{self}} = 
\begin{cases}
\mathbf{M}_t \cdot \mathbf{V}_t^*, & \text{if } t < \tau_s, \\
\softmax\left( 
\left[ 
f(\mathbf{Q}_t^*,\mathbf{K}_t^*) 
\;\middle|\; 
\frac{\mathbf{Q}_t^* \cdot \mathbf{K}_t^\top}{\sqrt{d}} \otimes \mathcal{M}^b 
\right]
\right) \cdot \mathbf{V}_t^{\prime}, 
& \text{otherwise}.
\end{cases}
\label{eq: self-attention}
\end{equation}}

The resulting attention map is then used to aggregate the corresponding values $\mathbf{V}_t^{\prime}$. We formulate the attention reweighting function $f(\cdot)$ as follows:
\begin{equation}
\begin{aligned}
f(\mathbf{Q}_t^*,\mathbf{K}_t^*) = \softmax\left(\frac {\mathbf{Q}_t^*(\mathbf{K}_t^*)^\top + \lambda \mathbf{M}^{\text{self}}_t}{\surd d}\right)
\end{aligned}
\label{eq:SA}
\end{equation}
Here, we apply the same modulation formula in Eq.~\eqref{eq:ca-bias} for self-attention, replacing cross-attention features with self-attention queries and keys to form $\mathbf{M}^{\text{self}}_t$. To ensure that spatial tokens only attend to semantically identical regions, we define the constraint matrix $\mathbf{C}$ for self-attention, which indicates whether two spatial tokens share the same instance identity across masks. This constraint is formally defined as:
\begin{equation}
\mathbf{C}[i,j] =
\begin{cases}
1, & \forall n \in \{1,\dots,N\},\; \mathcal{M}_n[i] = \mathcal{M}_n[j], \\
0, & \text{otherwise}.
\end{cases}
\end{equation}


\section{Experiments}
\label{sec:majhead}
\subsection{Experiment settings}
\label{ssec:ex_set}
\noindent\textbf{Datasets.} We evaluate our method on a curated benchmark of 75 videos at 512×512 resolution with 16 frames per video, comprising 25 single-object and 50 multi-object editing videos. Videos are sourced from DAVIS~\cite{davis}, YouTube-VOS~\cite{vis2022}, and public Internet sources. For Internet videos, object instance masks are generated using SAM3~\cite{sam3}. Video captions and editing prompts are either sourced from the respective datasets or generated via ChatGPT.

\noindent\textbf{Baselines.} We compare our method with four state-of-the-art text-driven video editing approaches: TokenFlow~\cite{tokenflow}, Ground-A-Video~\cite{gav}, VideoDirector~\cite{videodirector}, and VideoGrain~\cite{videograin}. We adopt the default implementations for all methods. For fair comparison, all ControlNet modules are disabled.

\begin{table*}[t]
\caption{Comparison of performance on single-object and multi-object video editing. Automatic evaluation results are reported as \textit{single-object score / multi-object score}. Best results are shown in \textbf{bold}, and second-best results are \underline{underlined}. VideoDirector with multi-pass editing is marked with (*), while the unmarked version refers to single-pass editing.}
\centering
\small 
\setlength{\tabcolsep}{4pt} 
\begin{tabular}{l c ccccc ccc}
\toprule
\textbf{Method} & \textbf{Venue} & \multicolumn{5}{c}{\textbf{Automatic Metrics (Single-object / Multi-object)}} & \multicolumn{3}{c}{\textbf{User Study}} \\

\cmidrule(r){3-7} \cmidrule(l){8-10}
& & SC $\uparrow$ & BC $\uparrow$ & MS $\uparrow$ & CLIP-F $\uparrow$ & CLIP-T $\uparrow$
& Edit-Acc $\uparrow$ & TC $\uparrow$ & Overall $\uparrow$ \\

\midrule
TokenFlow & ICLR'24
& 93.16/93.51 & 92.23/93.13 & 96.65/96.32 & 97.32/97.66 & 31.32/32.03
& 2.78 & \underline{3.28} & 2.99 \\

Ground-A-Video & ICLR'24
& 93.02/93.33 & 92.54/94.03 & 95.09/95.56 & 96.72/97.33 & \textbf{32.23}/\underline{34.43}
& 2.63 & 2.02 & 2.25 \\

VideoGrain & ICLR'25
& \underline{93.69}/\underline{94.21} & \underline{94.32}/\underline{94.81}
& \underline{96.79}/\underline{96.67} & \underline{97.43}/\underline{97.68} & 31.18/32.46
& 2.43 & 2.98 & 2.58 \\

VideoDirector & CVPR'25
& 93.39/86.85 & 93.61/90.73 & \textbf{97.12}/94.40 & 97.40/95.42 & 31.20/30.33 
& - & - & -\\

$\text{VideoDirector}^{*}$ & CVPR'25
& 93.39/93.72 & 93.61/94.34 & \textbf{97.12}/96.91 & 97.40/97.02 & 31.20/31.49
& \underline{2.89} & 2.83 & \underline{2.80} \\

\rowcolor{teal!20} \textbf{IDAG-Edit (Ours)} & -
& \textbf{94.13}/\textbf{94.31} & \textbf{94.70}/\textbf{94.83}
& \textbf{97.12}/\textbf{97.11} & \textbf{97.44}/\textbf{97.74}
& \underline{31.47}/\textbf{34.76}
& \textbf{3.28} & \textbf{3.34} & \textbf{3.23} \\

\bottomrule
\end{tabular}
\label{tab:single_multi_comparison}
\end{table*}

\subsection{Evaluation Results}
\label{ssec:Evaluation Results}
\noindent\textbf{Qualitative Comparison.} 
Fig.~\ref{fig:fig1} and Fig.~\ref{fig:qualitative} demonstrate that our method consistently outperforms prior approaches in multi-object editing and temporal consistency. In the second example of Fig.~\ref{fig:qualitative}, TokenFlow fails to correctly edit the horse and introduces background artifacts, while Ground-A-Video and VideoGrain only partially edit the rider and fail to transform the horse. Although VideoDirector follows the prompt, it exhibits severe color flickering across frames. In the last example, TokenFlow, Ground-A-Video, and VideoDirector suffer from attribute binding, and VideoGrain fails to preserve pose and appearance consistency. In contrast, IDAG-Edit enables well-localized edits while simultaneously preserves background fidelity and consistent object identities over time.


\noindent\textbf{Quantitative Comparison.}
We evaluate our method on both single- and multi-object video editing scenarios using the benchmark described at \cref{ssec:ex_set}. Following VBench~\cite{vbench++}, we report automatic metrics including Subject Consistency (SC), Background Consistency (BC), Motion Smoothness (MS), as well as CLIP-F and CLIP-T~\cite{clip} to assess perceptual consistency and text–prompt alignment. We also conduct a user study with 24 participants evaluating Edit Accuracy (Edit-Acc), Temporal Consistency (TC), and Overall Edit Quality (Overall), in which all methods are ranked from 1 (worst) to 5 (best). As shown in Table~\ref{tab:single_multi_comparison}, IDAG-Edit achieves improves object-level consistency and temporal coherence over existing methods, particularly in multi-object scenarios, reflecting stronger robustness against identity drift and cross-object attribute interference.

\begin{figure}[t]
    \centering
    \includegraphics[width=1.0\linewidth]{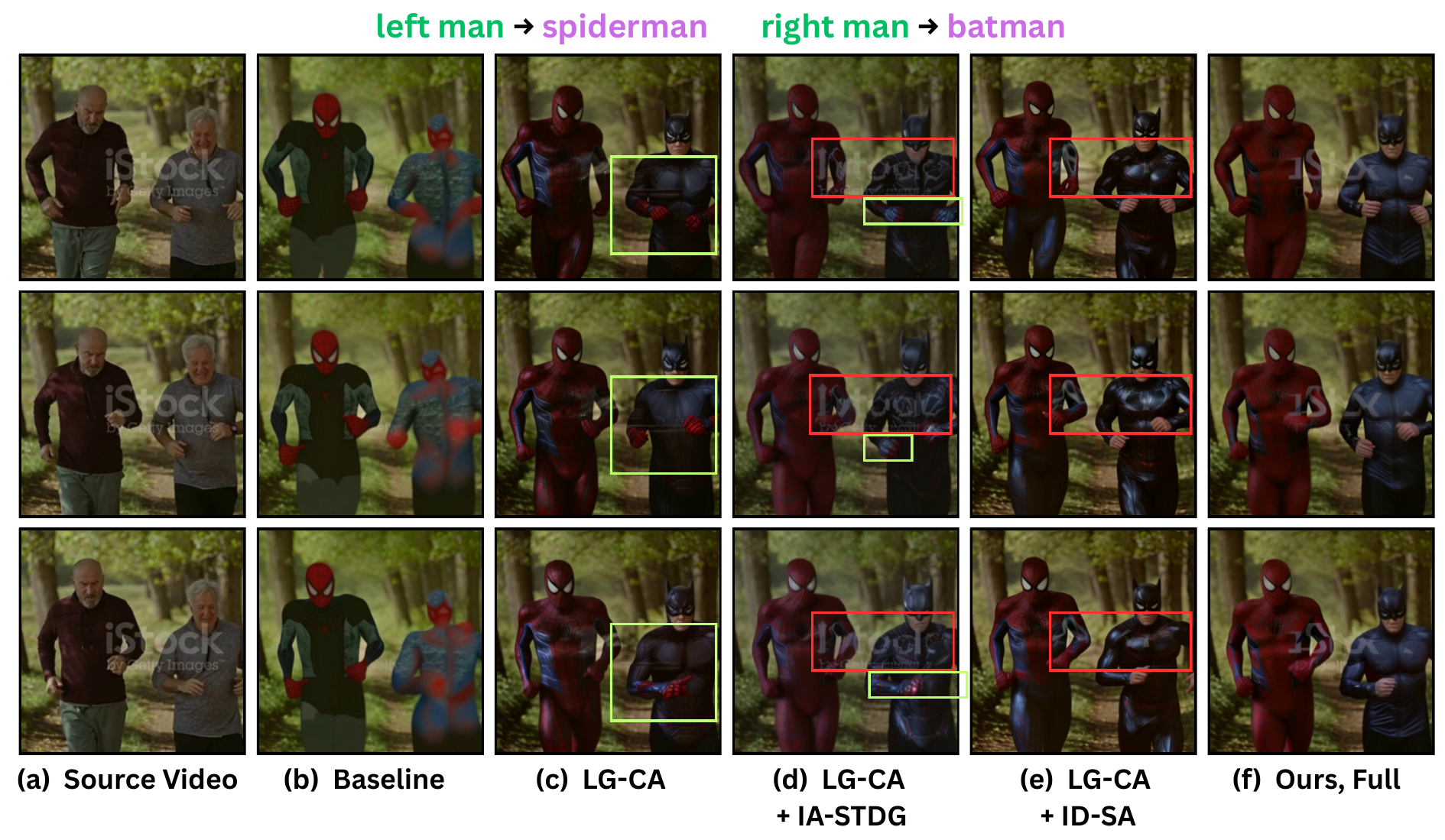}
    \caption{Ablation Study of LG-CA, IA-STDG and ID-SA.}
    \label{fig:ablation}
\end{figure}

\begin{table}[t]
    \caption{Quantitative ablation of IDAG-Edit components.}
    \centering
    \scriptsize 
    \setlength{\tabcolsep}{2pt}
    \begin{tabular}{c|ccc|ccccc}
    \toprule
    \multirow{2}{*}{\textbf{Settings}} & \multicolumn{3}{c|}{\textbf{Components}} & \multicolumn{5}{c}{\textbf{Multi-Object editing}} \\

    \cmidrule(r){2-4} \cmidrule(l){5-9}
    & LG-CA & IA-STDG & ID-SA & SC $\uparrow$ & BC $\uparrow$ & MS $\uparrow$ & CLIP-F $\uparrow$ & CLIP-T $\uparrow$ \\

    \midrule
    A & & &  & 86.85 & 90.73 & 94.40 & 95.42 & 30.33 \\
    B & $\checkmark$ & &  & 93.91 & 94.78 & 96.91 & 97.49 & 32.60 \\
    C & $\checkmark$ & $\checkmark$ & & 93.22 & 94.05 & 97.09 & 97.69 & 32.99 \\
    D & $\checkmark$ & & $\checkmark$  & 94.23 & 94.67 & 96.99 & 97.64 & 33.48 \\
    E & $\checkmark$ & $\checkmark$ & $\checkmark$ & \textbf{94.31}  & \textbf{94.83}  & \textbf{97.11}  & \textbf{97.74}  & \textbf{34.76} \\
    \bottomrule
    \end{tabular}
    \label{tab:ablation}
\end{table}


\subsection{Ablation Study}
\label{ssec:subhead}

\noindent\textbf{Layout-Guided Cross-Attention.}
As shown in Fig.~\ref{fig:ablation} and Tab.~\ref{tab:ablation}, the baseline (Setting A) fails to spatially distinguish between instances during editing. While LG-CA improves text-to-region alignment and enables localized edits (Setting B), it remains insufficient for fine-grained control, as visual attributes often leak across instance boundaries, leading to undesirable semantic contamination.


\noindent\textbf{Instance-Aware Spatial-Temporal Decoupled Guidance.}
The original STDG~\cite{videodirector} without instance awareness (Fig.~\ref{fig:ablation}-c,e) applies guidance using a unified mask formed by merging all object instances, resulting in coupled gradient updates across objects. This design suppresses instance-specific variations, yielding superficially higher consistency at the expense of editing flexibility and with increased cross-object interference. IA-STDG instead decouples instance-wise gradient fields, enabling independent and precise multi-object control (Setting C). However, strict spatial decoupling on entangled self-attention features induces feature–gradient mismatch, which can destabilize denoising dynamics and cause visual artifacts (e.g., hand distortions in Fig.~\ref{fig:ablation}-d), along with a slight performance drop in Tab.~\ref{tab:ablation}, row 3.

\noindent\textbf{Instance-Decoupled Self-Attention.}
To address the aforementioned feature entanglement, ID-SA restricts queries to instance-specific regions, effectively yielding clean object disentanglement. Although combining LG-CA with ID-SA (Setting D) achieves separation, it lacks the precise gradient supervision of IA-STDG, resulting in background degradation as shown in ~\cref{fig:ablation}e. By integrating all components (Setting E), IA-STDG and ID-SA provide complementary gradient-level and feature-level constraints, preserving background fidelity while maintaining instance separation, leading to the best overall consistency.

\section{conclusion}
\label{sec:conclusion}
In this paper, we propose a unified framework for multi-object video editing that enhances text-to-video diffusion models with instance-aware attention and guidance mechanisms. By integrating layout-guided cross-attention, instance-decoupled self-attention, and instance-aware spatial-temporal decoupled guidance, our method effectively mitigates attribute entanglement and semantic leakage while improving temporal consistency in complex scenes. Extensive experiments show that our approach outperforms prior methods in both single- and multi-object settings, achieving better prompt alignment, subject consistency, and motion smoothness.

\section{Acknowledgments}
This work was financially supported in part (project number: 112UA10019) by the Co-creation Platform of the Industry Academia Innovation School, NYCU, under the framework of the National Key Fields Industry-University Cooperation and Skilled Personnel Training Act, from the Ministry of Education (MOE) and industry partners in Taiwan. It also supported in part by the National Science and Technology Council, Taiwan, under Grant NSTC-115-2634-F-A49-011-, NSTC-114-2218-E-A49-024-, Grant NSTC-112-2221-E-A49-089-MY3, Grant NSTC-115-2425-H-A49-001, Grant NSTC-114-2622-E-A49-027, Grant NSTC-112-2221-E-A49-092-MY3, and in part by the Higher Education Sprout Project of the National Yang Ming Chiao Tung University and the Ministry of Education (MOE), Taiwan. It is also partly supported by MediaTek Inc., Hon Hai Research Institute, and Industrial Technology Research Institute.


\small
\bibliographystyle{IEEEbib}
\bibliography{refs}
\end{document}